\documentclass[11pt]{article}
\usepackage[margin=1.25in]{geometry}
\usepackage{microtype}

\usepackage{graphicx}
\usepackage{amsmath,amssymb,amsthm,mathtools}
\usepackage{bm}
\usepackage{booktabs}
\usepackage{listings}
\usepackage{subcaption}
\usepackage{float}
\usepackage{microtype}

\usepackage[numbers,sort&compress]{natbib}
\usepackage[hidelinks]{hyperref}

\title{Temporal Knowledge-Graph Memory in a Partially Observable Environment}

\author{
	Taewoon Kim\textsuperscript{\rm 1}\\
	\textsuperscript{\rm 1}HumemAI\\
	\texttt{taewoon@humem.ai}
	\and
	Vincent François-Lavet\textsuperscript{\rm 2}\\
	\textsuperscript{\rm 2}Vrije Universiteit Amsterdam\\
	\texttt{vincent.francoislavet@vu.nl}
	\and
	Michael Cochez\textsuperscript{\rm 3}\\
	\textsuperscript{\rm 3}ELLIS Institute Finland \& Abo Akademi University\\
	\texttt{Michael.cochez@abo.fi}
}

\begin{document}
\maketitle

\begin{abstract}
Agents in partially observable environments require persistent memory to integrate observations over time. While KGs (knowledge graphs) provide a natural representation for such evolving state, existing benchmarks rarely
expose agents to environments where both the world dynamics and the
    agent’s memory are explicitly graph-shaped. We introduce the Room Environment v3, a configurable environment whose hidden state is an RDF KG and whose observations are RDF triples.
The agent may extend these observations into a temporal KG when storing
them in long-term memory. The environment is easily adjustable in terms of grid
size, number of rooms, inner walls, and moving objects.

We define a lightweight temporal KG memory for agents, based on
RDF-star-style qualifiers (\texttt{time\_added},
\texttt{last\_accessed}, \texttt{num\_recalled}), and evaluate several
symbolic baselines that maintain and query this memory under different
capacity constraints. Two neural sequence models (LSTM and Transformer) serve as contrasting baselines without explicit KG structure. Agents train on one layout and are evaluated on a held-out layout with the same dynamics but a different query order, exposing train–test generalization gaps. In this setting, temporal qualifiers lead to more
stable performance, and the symbolic TKG (temporal knowledge graph) agent achieves roughly fourfold higher test QA (question-answer) accuracy than the neural baselines under the same environment and query conditions. The environment, agent implementations, and
experimental scripts are released for reproducible research at
\url{https://github.com/humemai/agent-room-env-v3} and \url{https://github.com/humemai/room-env}.

\end{abstract}

\section{Introduction}

Agents operating in partially observable environments must infer and maintain structured information about the world as it evolves over time~\cite{KAELBLING199899}. Because the world evolves according to structured dynamics, agents must store and update facts observed many steps earlier. Knowledge graphs (KGs)
provide a natural representational substrate for such settings: entities,
relations, and qualifiers can express spatial structure, object
locations, and temporal metadata in a uniform semantic framework~\cite{ehrlinger2016towards,Hogan_2021,battaglia2018relationalinductivebiasesdeep,zambaldi2018relationaldeepreinforcementlearning}. However, most benchmarks do not expose KG-shaped hidden states nor agents with explicitly modeled temporal KG memory (e.g., \cite{battaglia2018relationalinductivebiasesdeep,zambaldi2018relationaldeepreinforcementlearning,Cote2018TextWorld,Hausknecht2020Jericho,Ammanabrolu2020Graph,Wayne2018MERLIN,graves2014neural,graves2016hybrid}).

This paper introduces the Room Environment v3, a deterministic and fully configurable
environment designed to fill this gap. The environment’s
hidden state is an RDF KG whose entities include rooms,
objects, walls, and the agent, and whose relations encode spatial
adjacency and object locations. At each timestep the agent receives a
symbolic observation consisting of RDF triples describing the local room
structure and the objects currently present. The environment is easily
adjustable in terms of grid size, number of rooms, pattern of inner
walls, and the number and motion patterns of moving objects. A further
component of the environment is a query at each step requesting the
current location of a named object. Accurate responses require
maintaining an internal representation of the world that integrates
partial observations over time.

While KGs and temporal KGs are widely used for representing evolving
knowledge~\cite{Hogan_2021,dellaglio2017streamreasoning,moreau2012prov,DBLP:journals/corr/abs-1809-03202},
their application as \emph{agent memory} in interactive environments
remains underexplored. Prior work on gridworlds, symbolic environments,
and relational reasoning benchmarks typically either (i) does not
provide a KG-shaped hidden state, (ii) gives agents unstructured
symbolic or text observations without a defined KG memory model, or
(iii) uses neural representations that do not expose explicit triples or
qualifiers for inspection~\cite{Cote2018TextWorld,Hausknecht2020Jericho,Ammanabrolu2020Graph,10.1609/aaai.v37i1.25075,battaglia2018relationalinductivebiasesdeep,zambaldi2018relationaldeepreinforcementlearning,graves2014neural,graves2016hybrid,Wayne2018MERLIN,DBLP:journals/corr/abs-1803-10122}.
 As a result, there is limited
understanding of how explicit temporal KGs function as internal state
representations, how simple symbolic update rules perform relative to
neural sequence models, and how temporal qualifiers influence
generalization.

Recent work on long-term memory in artificial intelligence has
increasingly focused on neural vector-based representations, where
stored information is encoded in continuous embeddings rather than
explicit symbolic structures~\cite{graves2014neural,graves2016hybrid,Wayne2018MERLIN,DBLP:journals/corr/abs-1803-10122,pritzel2017neuralepisodiccontrol,blundell2016modelfreeepisodiccontrol}. While such
approaches are powerful, they often lack interpretability: the contents
of memory and the basis of retrieved answers are not directly observable
as explicit symbolic facts. One of the motivations for our contribution
is to provide a setting where both the hidden state of the world and the
agent’s long-term memory are represented as explicit KGs. This enables interpretable memory inspection, deterministic updates, and transparent reasoning.

We propose a lightweight temporal KG memory model for agents that
extends standard RDF triples with RDF-star-style qualifiers capturing
\texttt{time\_added}, \texttt{last\_accessed}, and
\texttt{num\_recalled}. These qualifiers provide a simple but expressive
mechanism for tracking the recency and usage of stored facts, enabling
memory strategies that remain fully interpretable~\cite{10.1145/2452376.2452478}. Based on this TKG
framework, we implement several symbolic baselines that update and query
their memory using deterministic rules, with optional capacity limits
and simple eviction heuristics. To provide a contrasting perspective, we
include two neural sequence models (LSTM and Transformer) that receive exactly the same symbolic observations but maintain memory as a fixed-length sequence
queue buffer without explicit KG structure.

We evaluate all agents on two deterministic layouts: a training
environment and a held-out test environment with identical dynamics but
a different question order. This setup enables a controlled examination
of train-to-test generalization and highlights the importance of
structured temporal memory. Across capacities, temporal qualifiers
contribute to more stable behavior, and the symbolic TKG agent exhibits
significantly higher test performance than the neural baselines under the same environment and query conditions. The environment, agent implementations,
and full reproducibility material are publicly released in anonymized
form.

\subsection*{Contributions}

\begin{itemize}
    \item We introduce the Room Environment v3, a deterministic and configurable environment
        whose hidden state and observations are expressed as RDF knowledge graphs.
    \item We define a lightweight temporal KG memory with RDF-star-style
    qualifiers and deterministic update rules.
    \item We compare symbolic TKG agents with neural sequence baselines
    under a shared interface and varying memory capacities.
    \item We release the full environment, agent implementations, and
    reproducibility scripts.
\end{itemize}

\paragraph{Code and environment.}
The main repository is at \url{https://github.com/humemai/agent-room-env-v3};
the Room Environment repository is at \url{https://github.com/humemai/room-env}.

\subsection*{Version history}
This paper is a substantially revised and extended version of an earlier
arXiv preprint on KG-based memory for partially observable environments.
The present version introduces a new deterministic Room Environment v3,
reformulates agent memory explicitly as a temporal knowledge graph using
RDF-star qualifiers, adds symbolic baselines with interpretable update and
eviction rules, and provides a controlled generalization analysis against
neural sequence models. The earlier formulation is subsumed by this work,
and citations to prior versions should be understood as referring to the
present manuscript.

\section{Background}

\subsection{RDF, RDF-star, and Knowledge Graph Representations}

The Resource Description Framework (RDF) models facts as triples
$(s,p,o)$ with subject, predicate, and object drawn from IRIs or
literals. An RDF graph therefore corresponds to a directed,
edge-labeled multigraph in which relations capture structured
connections between entities. This representation is well suited for
spatially structured domains: rooms, objects, and adjacency relations
can be expressed directly as triples without task-specific encodings.

RDF-star extends RDF with a concise syntax for \emph{statements about
statements}. An embedded triple $\ll s\ p\ o\gg$ can be used as the
subject of further triples, e.g.,
$\ll s\ p\ o\gg\ \texttt{:time\_added}\ 7$. This allows qualifiers
(timestamps, counters, provenance keys) to be attached directly to a
base fact without the heavier reification vocabulary~\cite{hartig2021foundationsalternativeapproachreification}. Because embedded
triples behave like ordinary RDF terms, RDF-star offers a lightweight
and standard-compatible way to represent temporal annotations in an
agent’s long-term memory.

\subsection{Temporal Knowledge Graphs}

Temporal knowledge graphs (TKGs) represent facts whose validity or
relevance changes over time. Common approaches include interval-based,
event-based, and update-based models, all of which associate each
triple with temporal metadata reflecting when it was introduced or last
observed. RDF-star can express such metadata by attaching qualifier
key–value pairs to embedded triples, enabling fine-grained annotations
such as \texttt{time\_added}, \texttt{last\_accessed}, or
\texttt{num\_recalled}~\cite{moreau2012prov}. Temporal KGs have been widely used in the
Semantic Web to model evolving data, streaming updates~\cite{dellaglio2017streamreasoning}, and temporally
indexed knowledge sources.

In interactive settings, temporal qualifiers allow agents to track the
recency and usage of stored information, to maintain evolving internal
models, and to reason over past observations. These properties align
naturally with the needs of partially observable environments, where an
agent’s internal state must integrate information gathered over many
timesteps.

\subsection{Symbolic Memory and Semantic State Tracking}

Research on long-term memory for artificial agents has increasingly
explored neural vector-based representations, where stored information
is encoded implicitly in embeddings or hidden states~\cite{graves2014neural,graves2016hybrid,Wayne2018MERLIN,DBLP:journals/corr/abs-1803-10122,pritzel2017neuralepisodiccontrol,blundell2016modelfreeepisodiccontrol}. While effective for high-dimensional signals, such representations are typically opaque:
the contents of memory cannot be inspected directly as noted in neural-symbolic systems~\cite{besold2017neuralsymboliclearningreasoningsurvey}, and the basis for
retrieval is difficult to trace.

In contrast, symbolic memory systems store explicit facts that can be
examined, queried, and updated deterministically. KGs offer
a particularly suitable substrate for this purpose, as they represent
entities and relations at a semantic level and support principled
operations for insertion, deletion, and reannotation. When temporal
qualifiers are incorporated, symbolic memory becomes fully transparent:
each fact carries a clear history of when and how it was observed.

Heuristic update strategies such as recency-based selection,
frequency-based selection, and simple eviction rules (e.g.~FIFO, LRU,
LFU) can operate directly over qualifiers attached to embedded triples.
Such heuristics preserve interpretability while providing meaningful
signals for deciding which facts to retain when memory capacity is
limited.

\subsection{Semantic Representations for Agents and Interactive Environments}

Semantic Web research has long explored structured representations,
logical reasoning, and knowledge-driven decision processes. Several
interactive or agent-based settings use symbolic descriptions of the
world, but the underlying environments are rarely expressed directly as
KGs, and the agent’s internal memory is typically not a
temporal KG. Existing gridworld-style or relational reasoning benchmarks
often provide symbolic observations without specifying how those
observations should be integrated into a structured, evolving memory.

As a result, there is limited empirical understanding of how temporal
KGs function as internal state representations for agents, how simple
qualifier-driven heuristics compare to unstructured sequence-based
memory, and how temporal metadata affects generalization in deterministic
but partially observable domains. The environment and memory model
introduced in this paper address these gaps by providing a setting in
which both the hidden state of the world and the agent’s memory are
explicit, inspectable knowledge graphs represented in standard RDF and
RDF-star syntax.

\section{Room Environment v3}
\label{sec:env}

\begin{figure}[H]
\centering
\includegraphics[width=\linewidth]{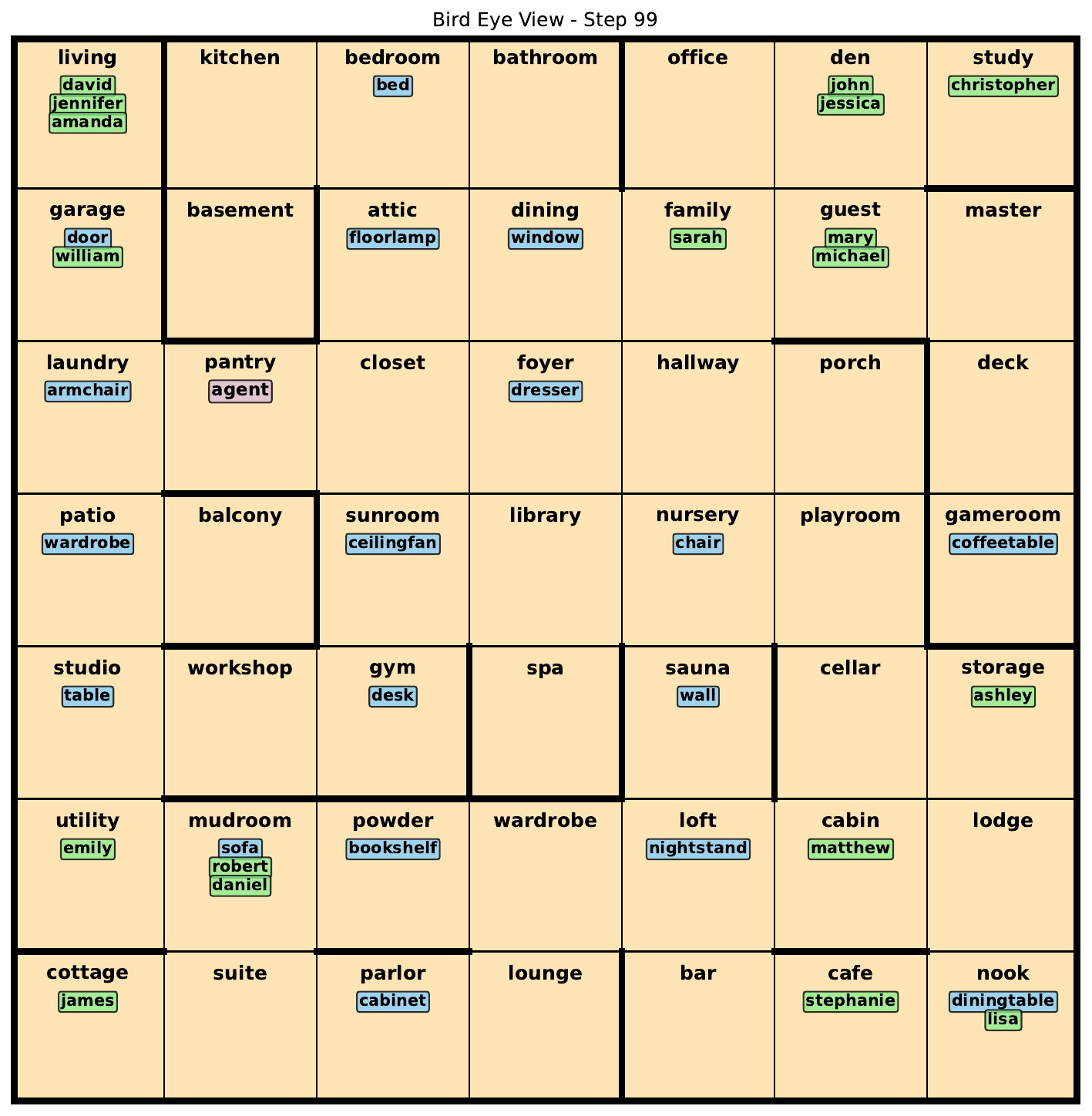}
\caption{Bird's-eye schematic of the hidden state at $t=99$ ($s_{t=99}$),
showing spatial layout and entity placement. This view is only a schematic for intuition:
the actual environment state and agent-facing world are represented in the RDF knowledge-graph
world (Figure~\ref{fig:hidden-state-rdf-step99}). The agent does not directly observe $s_t$;
its observation $o_t$ is the induced RDF subgraph of the current room and visible adjacency
relations.}
\label{fig:bird-eye-view-step99}
\end{figure}

\begin{figure}[H]
\centering
\includegraphics[width=\linewidth]{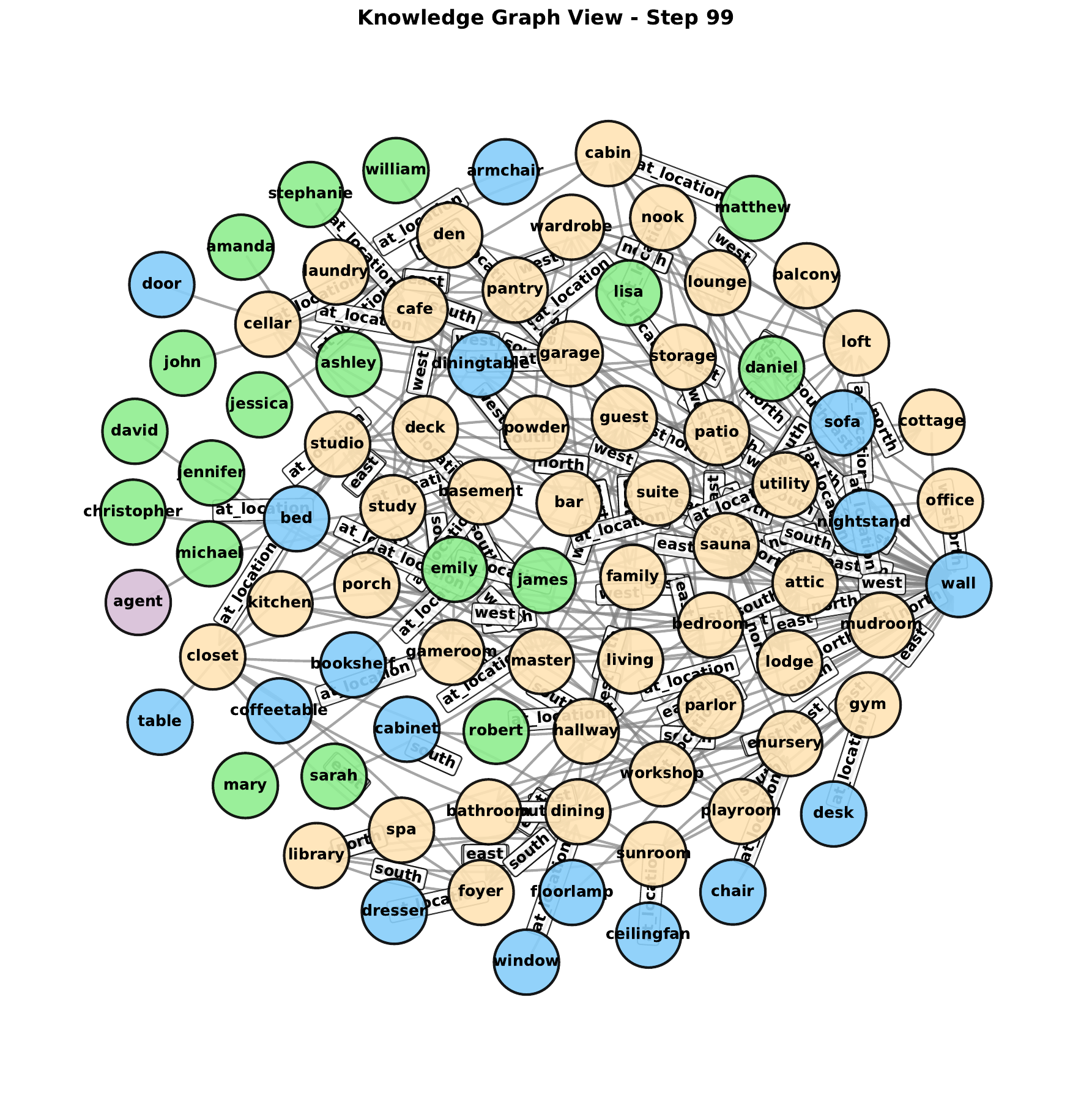}
\caption{Knowledge-graph view of the same hidden state at $t=99$ ($s_{t=99}$), expressed
as RDF structure.}
\label{fig:hidden-state-rdf-step99}
\end{figure}

The Room Environment v3 (Figures~\ref{fig:bird-eye-view-step99} and
\ref{fig:hidden-state-rdf-step99}) is a configurable
$g \!\times\! g$ grid world populated with named rooms, static and moving
objects, and periodically changing inner walls. All environment components—hidden dynamics, observations, and the QA loop—are defined in RDF triples triples, making the environment a minimal but fully KG-shaped testbed for studying temporal memory.

The environment deliberately couples navigation and question answering. Because objects move and visibility is local, correct answers require exploration, accumulation of partial observations, and persistent memory of object trajectories. This design prevents trivial solutions: without navigation the task reduces to static KG lookup, and without queries exploration has no semantic purpose. Their combination creates a controlled setting in which temporal memory is both necessary and measurable.

The environment generalizes the earlier setting introduced by
Kim~et~al.~\cite{10.1609/aaai.v37i1.25075}: instead of a single fixed layout,
our version supports arbitrary grid sizes, heterogeneous wall schedules,
multiple moving objects with independent motion rules, and exposes the entire hidden state as an RDF KG; when used as long-term memory inside the agent, this can be extended into a temporal KG via qualifiers.

\subsection{Hidden State and Dynamics}

At timestep $t$, the hidden state $s_t$ consists of the on/off status of all inner walls, the positions of all static and moving objects, and the agent's current room.

Each room is a node, and cardinal directions are modeled as labeled edges
(\texttt{north}, \texttt{east}, \texttt{south}, \texttt{west}). Walls are
represented explicitly—e.g., \texttt{(room, north, wall)}—rather than by removing
adjacency edges, allowing the environment to expose both potential adjacencies
and their blocked status. Every moving object is equipped with a
fixed, ordered preference list drawn from
$\{\texttt{north},\texttt{east},\texttt{south},\texttt{west},\texttt{stay}\}$.
At each step it executes the \emph{first feasible} move in its list;
if none are feasible, it remains in place. This rule, combined with wall
schedules, makes object motion \emph{deterministic}.

The inner walls follow fixed periodic schedules: if inner wall $i$ has a
pattern of length $|p_i|$, then the joint wall configuration repeats every
$P_w = \operatorname{lcm}(|p_1|,\dots,|p_W|)$ steps. Moving objects induce an
additional period $P_m$. After an initial transient $T_{\text{settle}}$, the
global state obeys
\[
  s_{t+P} = s_t, \qquad P = \operatorname{lcm}(P_w,\, P_m),
\]
so the environment is a finite deterministic system with a compact periodic
orbit.

Although deterministic, the state space is combinatorially large. With
$R = g^2$ rooms, $M$ moving objects, and $W$ inner walls with pattern lengths
$|p_1|,\dots,|p_W|$, a coarse upper bound on the state space is
\[
  |\mathcal{S}| \;\le\; R^M \prod_{i=1}^{W} |p_i|.
\]
For our running configuration ($g{=}7$,
$R{=}49$, $M{=}18$, $W{=}36$), this exceeds $6.6\times 10^{33}$.
Exact planning is infeasible; agents must instead reason over their own growing memory $M_t$.

The periodic wall schedule and deterministic object motion are intentional
design choices: they ensure that the hidden state evolves in a structured but
nontrivial way, making it impossible for an agent to answer location queries
from the current observation alone. Because objects move and visibility is
local, accurate question answering requires integrating partial observations
over many steps. Periodicity keeps the environment fully deterministic and
replayable while still enforcing genuine long-term memory demands.

A detailed description of the periodicity analysis, including closed-form
bounds and deterministic replay examples, is provided in the public
anonymous repository.

\subsection{Observations as RDF Graph Fragments}

At timestep $t$, the agent receives an observation $o_t$ consisting of the RDF
subgraph induced by its current room: the room node for the agent’s location,
any outgoing direction-labeled edges to adjacent rooms that are not blocked by
walls, and object–location triples for all objects currently in that room. This
yields 5–6 triples per step. Observations contain no information
beyond the agent’s immediate neighborhood, so long-horizon queries require
integrating partial observations over time.

Each timestep also issues a \emph{location query} of the form: $(X,\ \texttt{:at\_location},\ ?)$.
The agent must identify the current room of the queried object~\cite{bordes2015largescalesimplequestionanswering}. This
requires maintaining a temporal model of object motion as observed
indirectly through partial glimpses, echoing aspects of KG-based QA (question-answer)~\cite{yih-etal-2015-semantic}.

\subsection{Deterministic Question–Move Loop}

At each timestep the agent first answers the query by returning the room of the
specified object (reward $+1$ if correct, otherwise $0$), and then executes a
movement action chosen from \{\texttt{north}, \texttt{east}, \texttt{south},
\texttt{west}, \texttt{stay}\}.

Episodes last 100 steps, a length chosen to balance tractability with sufficient temporal depth.
The full query sequence for each episode is pre-generated and fixed across all agents, ensuring strict comparability in evaluation.

\subsection{Configurable Layouts and Test Split}

The environment is fully parameterized by the grid length $g$, the number and
placement of static and moving objects, the number and periodic patterns of
inner walls, and the episode length.

We use two deterministic layouts: a training layout with fixed walls and object
preferences, and a held-out test layout with identical dynamics but a different
query order. This difference in question order yields a controlled train–test
generalization setting in which agents must transfer their learned memory
strategy rather than memorize a fixed query sequence.

\subsection{Memory Interface (Agent-Agnostic)}

The environment maintains only the hidden state $s_t$ and the symbolic
observation $o_t$ emitted at each timestep. It does not prescribe any internal
memory structure: agents may keep any private state they choose. The
environment simply provides RDF observations, receives an action and an answer,
and evaluates correctness against $s_t$.

In our experiments, we compare four agents. The \emph{RDF-star agent}
stores each observed fact as an RDF-star triple annotated with qualifiers
(\texttt{time\_added}, \texttt{last\_accessed}, \texttt{num\_recalled}). The
\emph{RDF agent} stores the same observations as plain RDF triples without
qualifiers. The two \emph{neural baselines} maintain fixed-length buffers of
tokenized observations without explicit KG structure; one uses an LSTM
encoder and the other uses a Transformer encoder. All agents operate in the same environment and query conditions.

\subsection{Default Configuration}

Unless stated otherwise, experiments use a \texttt{grid\_length} of $7$, with
$18$ static objects, $18$ moving objects, $36$ inner walls following periodic
patterns, an episode length of $100$, and a fixed query list shared across all
agents. The reward is simply the number of correct queries (maximum $100$).
All environment code, layouts, and deterministic replay scripts are included in
the public anonymous repository for full reproducibility.

\section{Agents}
\label{sec:agents}

We evaluate four agents: two \emph{symbolic} agents that store explicit
KG structures, and two \emph{neural} agents that store tokenized observation histories. All agents receive the same query sequence and operate in the same environment, but their exploration behaviors cause them to generate different observation histories. We evaluate them under memory capacities from 0 to 512 entries, which covers the range where capacity meaningfully constrains what can be stored. Table~\ref{tab:agents} summarizes the four agents.

\begin{table}[tb]
\centering
\small
\begin{tabular}{lcccc}
\toprule
\textbf{Agent} &
\textbf{Memory repr.} &
\textbf{Qualifiers} &
\textbf{QA rule} &
\textbf{Eviction} \\
\midrule
Symbolic--RDF &
RDF triples &
No &
SPARQL &
Random \\

Symbolic--TKG &
RDF-star (TKG) &
Yes &
MRA/MRU/MFU &
FIFO/LRU/LFU \\

Neural--LSTM &
Obs.\ history &
No &
Prediction head &
FIFO \\

Neural--Transformer &
Obs.\ history &
No &
Prediction head &
FIFO \\
\bottomrule
\end{tabular}
\caption{Agent comparison. MRA = most recently added, MRU = most recently used,
MFU = most frequently used.}
\label{tab:agents}
\end{table}

\subsection{Symbolic agents}

Both symbolic agents store explicit triples. Their difference is whether
they attach temporal qualifiers.

\paragraph{RDF agent (plain triples).}
Memory is a bounded set of RDF triples $(s,p,o)$.
Because these entries carry \emph{no} timestamps or usage counts, the
agent has no basis for ordering them. Therefore, when memory is full, the
only principled choice is to evict a \emph{uniformly random} triple.

\emph{QA rule.} Answers are obtained by plain SPARQL pattern matching:
\begin{lstlisting}[language=SPARQL,basicstyle=\ttfamily\small,frame=single]
SELECT ?o WHERE { <s> <r> ?o . }
ORDER BY RAND() LIMIT 1
\end{lstlisting}
If multiple candidates match, tie-breaking is uniform because the agent
lacks temporal metadata.

\emph{Exploration.}
The current map is reconstructed from the triples stored so far.
The agent performs BFS (breadth-first search) to the nearest unvisited room. BFS is used because the memory itself is a graph and BFS is the simplest deterministic
traversal aligned with that structure.

\paragraph{TKG agent (RDF-star with qualifiers).}
Each observed triple $(s,r,o)$ is stored as an embedded triple
$\ll s\ r\ o\gg$ annotated with:
\[
\texttt{time\_added},\quad \texttt{last\_accessed},\quad
\texttt{num\_recalled}.
\]
This creates a temporal knowledge graph $M_t$.

\emph{QA rules.}  
Each query $(s_q,r_q,?)$ is answered by selecting the object $o$ whose
qualifier is maximal:
\[
\begin{aligned}
\text{MRA} &: \arg\max \texttt{time\_added},\\
\text{MRU} &: \arg\max \texttt{last\_accessed},\\
\text{MFU} &: \arg\max \texttt{num\_recalled}.
\end{aligned}
\]

\emph{Eviction.}  
Using qualifiers permits structured removal:
FIFO (oldest \texttt{time\_added}),  
LRU (least recent \texttt{last\_accessed}),  
LFU (smallest \texttt{num\_recalled}).

\emph{Exploration.}  
The agent ranks frontier rooms using MRA/MRU/MFU on edge triples and
performs BFS to the top-ranked frontier.

\emph{SPARQL-star example (MRU).}
\begin{lstlisting}[language=SPARQL,basicstyle=\ttfamily\small,frame=single]
SELECT ?o WHERE {
  << <s> <r> ?o >> :last_accessed ?t .
}
ORDER BY DESC(?t)
LIMIT 1
\end{lstlisting}

Thus, the TKG agent is entirely deterministic and fully traceable: every
decision (answer, movement, eviction) is reproducible from $M_t$.

\subsection{Neural agents}

The neural agents store a fixed-length \emph{tokenized observation
history} without explicit graph structure. They differ only in the
sequence encoder used.

\paragraph{Joint decision structure.}
Unlike the symbolic agents, which decompose question answering and exploration into two independent rules, the neural agents produce a \emph{single joint operation} that simultaneously selects the answer and the movement. This results in a large option set (49 possible room answers $\times$ 5 movement choices) from which the prediction head chooses one element. Because answers referring to rooms that have not yet been observed are invalid, the option set is dynamically \emph{masked} at each timestep so that only admissible answers remain. Consequently, the neural agents must learn end-to-end how to resolve queries and how to navigate, without access to explicit graph structure, qualifiers, or symbolic update rules. The neural agents learn this joint policy using a standard reinforcement-learning objective implemented via a Deep Q-Network (DQN)~\cite{mnih2013playingatarideepreinforcement}.

\paragraph{Tokenization.}
Each triple $(s,p,o)$ is converted to an embedding vector
\[
\mathbf{e}_i
   = W_{\mathrm{emb}}\,[\,\mathrm{emb}(s)\;\Vert\;\mathrm{emb}(p)
      \;\Vert\;\mathrm{emb}(o)\,],
\]
and the last $L$ such tokens form the memory buffer. When full, the oldest token is removed (FIFO), reflecting the queue structure of a fixed-length sequence buffer.

\subsection*{LSTM neural agent}
The sequence $\mathbf{E}=[\mathbf{e}_1,\dots,\mathbf{e}_L]$ is encoded by:
\[
\mathbf{H} = \text{LSTM}(\mathbf{E}).
\]
Because LSTMs process data sequentially, temporal order is included
implicitly~\cite{6795963}.

\paragraph{Question-conditioned pooling.}
Given a question triple $(s_q,r_q,?)$, embed it as $\mathbf{q}$ and compute
attention weights:
\[
\alpha_i = \frac{\exp(\mathbf{q}^\top \mathbf{h}_i)}
{\sum_j \exp(\mathbf{q}^\top \mathbf{h}_j)},\qquad
\mathbf{c} = \sum_i \alpha_i \mathbf{h}_i.
\]
A feedforward prediction head maps $\mathbf{c}$ to the 245 possible
operation options (answer $\times$ movement).

\subsection*{Transformer neural agent}
Here, $\mathbf{E}$ is fed to a Transformer encoder:
\[
\mathbf{H} = \text{Transformer}(\mathbf{E} + \mathrm{PE}),
\]
where $\mathrm{PE}$ are sinusoidal positional encodings that supply the
temporal order~\cite{vaswani2023attentionneed}, enabling long-range temporal reasoning~\cite{dehghani2019universaltransformers}. The same question-conditioned pooling and prediction head
are used as in the LSTM agent.

\paragraph{Interpretability.}
Neural agents have no explicit triples, qualifiers, or traceable update
rules, similar to episodic-control agents~\cite{pritzel2017neuralepisodiccontrol,blundell2016modelfreeepisodiccontrol}. Their internal state is an opaque vector, and only soft attention
weights give partial insight into which past observations were influential.
Symbolic agents, in contrast, expose every stored fact and every eviction
decision.

\section{Experiments}

We evaluate all agents in the deterministic Room Environment v3 under varying long-term memory capacities, from 0 to 512. Each configuration is trained and tested on $5$ random seeds, and we report the mean and standard deviation. All raw results, hardware specifications, and hyperparameters are provided in the anonymous repository.

Our evaluation focuses on three aspects:

\begin{enumerate}
    \item \textbf{QA accuracy across memory capacities} (Figure~\ref{fig:qa_accuracy})
    \item \textbf{Coverage metrics over time} (Figure~\ref{fig:coverage})
    \item \textbf{Memory-state evolution for RDF-star} (Figure~\ref{fig:memory_states})
\end{enumerate}

\subsection{QA Accuracy Across Long-Term Memory Capacities}

\begin{figure}[tb]
    \centering
    \includegraphics[width=0.8\linewidth]{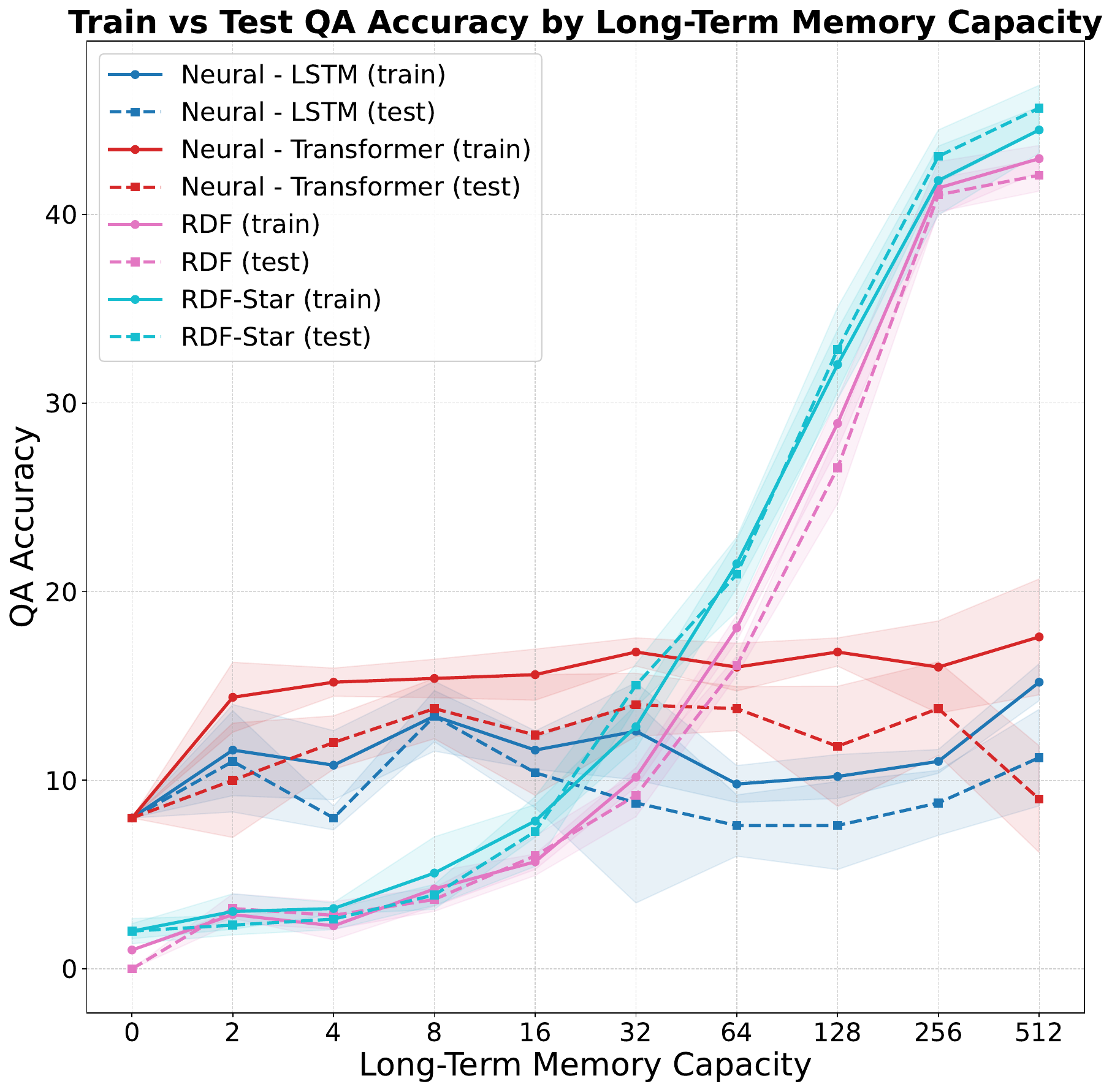}
    \caption{Train--test QA accuracy for all agents across long-term memory capacities. Mean and standard deviation are computed across $5$ seeds. Raw values can be found in the anonymous repository.}
    \label{fig:qa_accuracy}
\end{figure}

Figure~\ref{fig:qa_accuracy} shows the train and test QA accuracy for all agents.
Neural agents learn exploration and question answering jointly under a single end-to-end policy, yielding a large discrete action space and difficult credit assignment. As memory capacity increases, these agents also receive longer observation histories, making the underlying learning problem more computationally demanding; in principle, higher-capacity models may require substantially more training to benefit from the additional information. In our experiments, however, all memory-capacity settings were trained for the same number of episodes (200) to ensure a fair comparison, which likely contributes to the minimal performance gains observed in the higher-capacity regime.

Across almost all capacities, the Transformer agent slightly outperforms the LSTM agent, consistent with empirical trends in sequence modeling. However, neither architecture meaningfully exploits larger memory, and both exhibit substantial train--test gaps. In low-memory regimes ($0$--$16$), neural agents outperform symbolic ones---but this reflects random trial-and-error rather than informed reasoning: the agents often guess object locations without provenance and only gradually reinforce correct answers through reward.

Symbolic agents begin to outperform neural agents at capacity $32$. Their
accuracy increases with memory because operations such as BFS and graph queries
add computation but not statistical difficulty. Neural agents, by contrast, must
learn these structures from data and therefore require much more training and
capacity to benefit from larger memory.

RDF-star achieves better QA accuracy in most memory capacities. The RDF-star
agent family consists of 27 variants, arising from all combinations of three
qualifier-based QA rules (MRA, MRU, MFU), three exploration priorities based on
the same qualifiers, and three eviction heuristics (FIFO, LRU, LFU); these
choices are independent because each stored fact carries the qualifiers
\texttt{time\_added}, \texttt{last\_accessed}, and \texttt{num\_recalled}. We
report the best-performing variant, selected by training accuracy and evaluated
identically at test time. Although the plain RDF agent can store more main
triples under a fixed memory budget—since it stores triples rather than
quadruples—the temporal qualifiers in RDF-star yield a more informative and
expressive memory state, leading to stronger performance once capacity is
sufficient to retain both facts and their associated temporal metadata.

At capacity $512$, the best symbolic agent (RDF-star) reaches a QA accuracy of $46.52$, whereas the best neural agent (LSTM) reaches $11.2$, a four-fold difference. These results highlight that explicit symbolic memory is considerably more effective for long-horizon, partially observable semantic reasoning. Statistical variation is shown as shaded regions.

\subsection{Coverage Metrics: Room and Triple Coverage}

\begin{figure}[tb]
    \centering

    \begin{minipage}{0.49\linewidth}
        \centering
        \includegraphics[width=\linewidth]{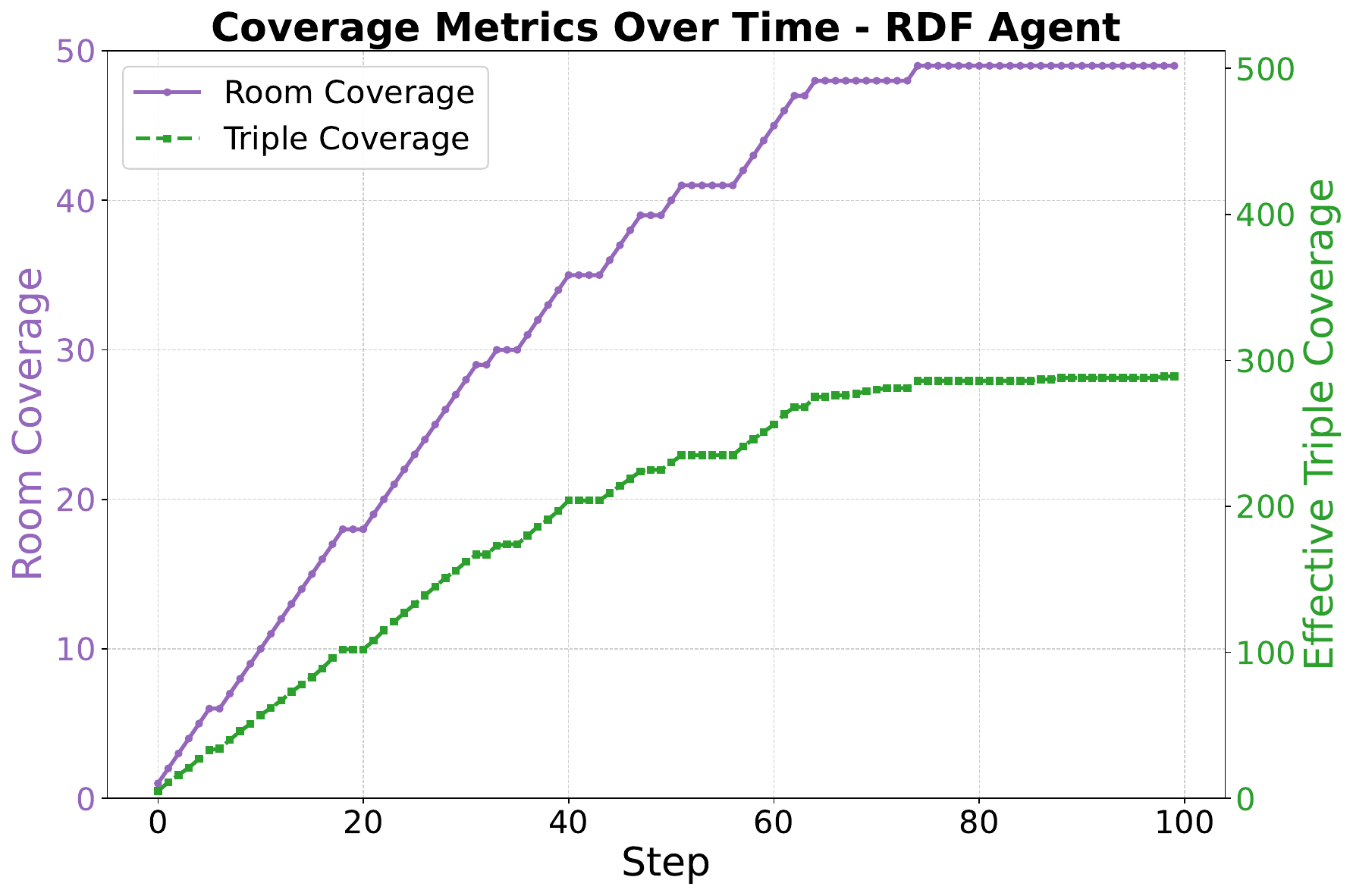}
        \textbf{(c) RDF}
    \end{minipage}
    \hfill
    \begin{minipage}{0.49\linewidth}
        \centering
        \includegraphics[width=\linewidth]{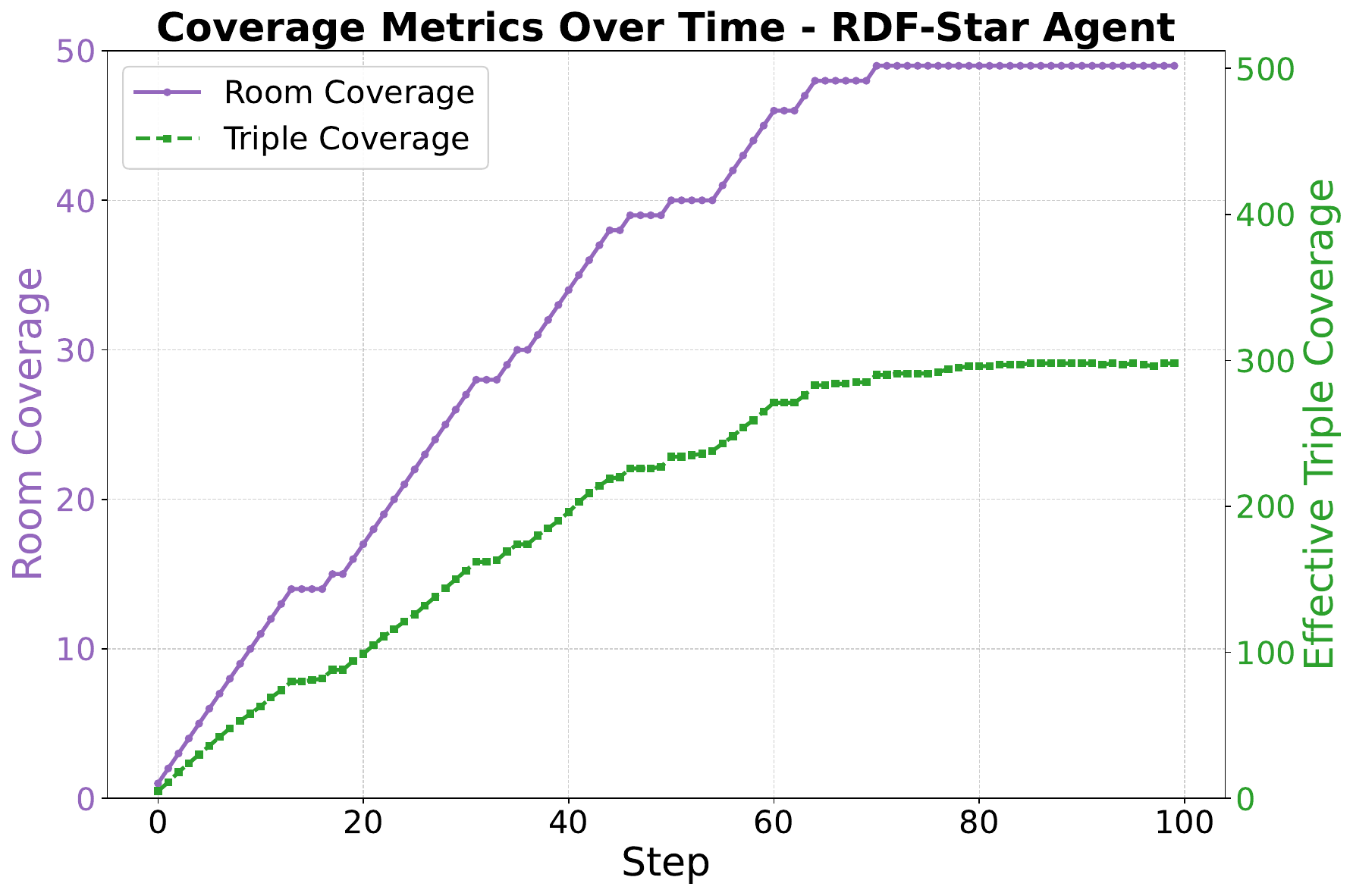}
        \textbf{(d) RDF-Star}
    \end{minipage}

    \begin{minipage}{0.49\linewidth}
        \centering
        \includegraphics[width=\linewidth]{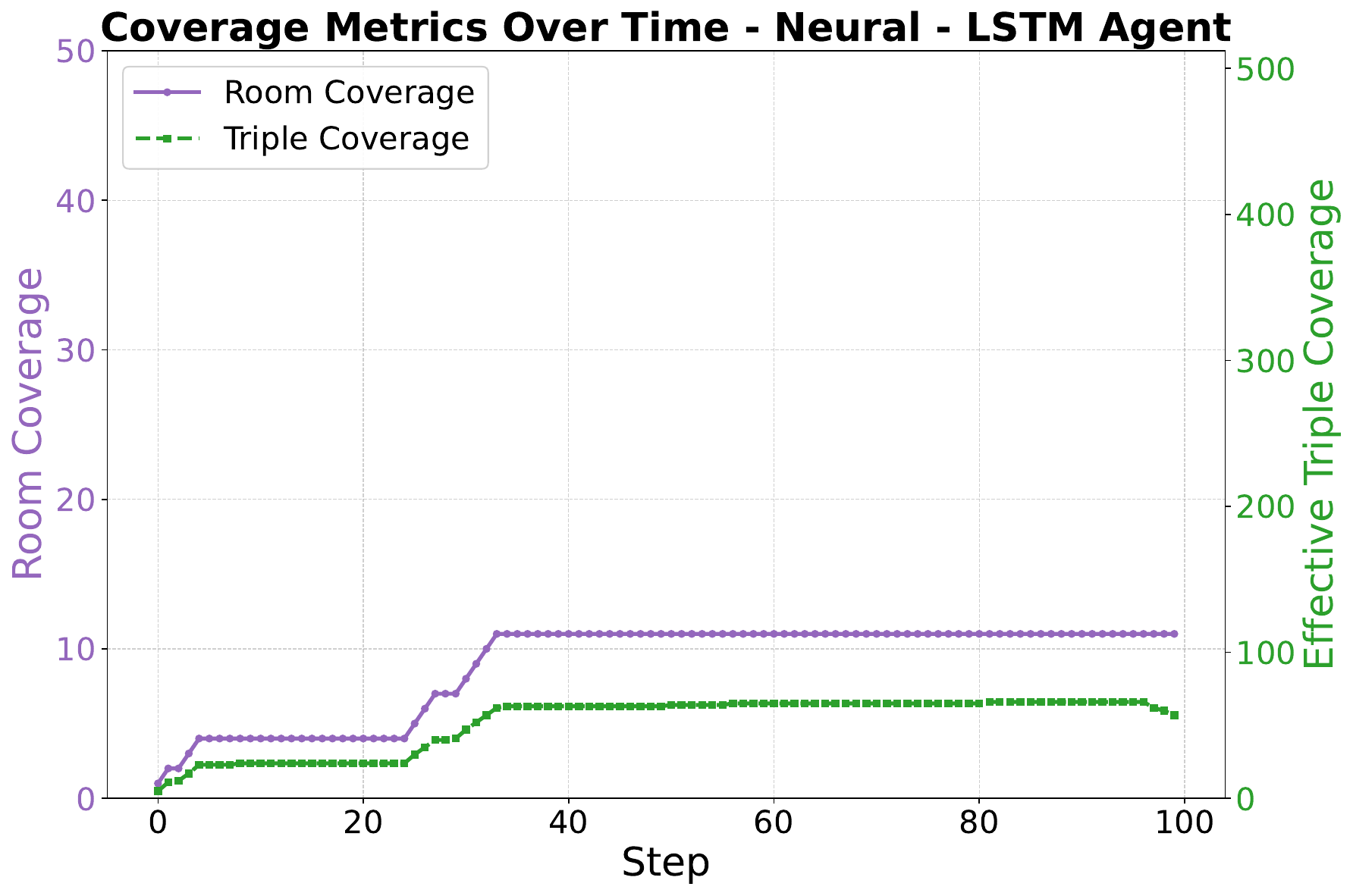}
        \textbf{(a) Neural -- LSTM}
    \end{minipage}
    \hfill
    \begin{minipage}{0.49\linewidth}
        \centering
        \includegraphics[width=\linewidth]{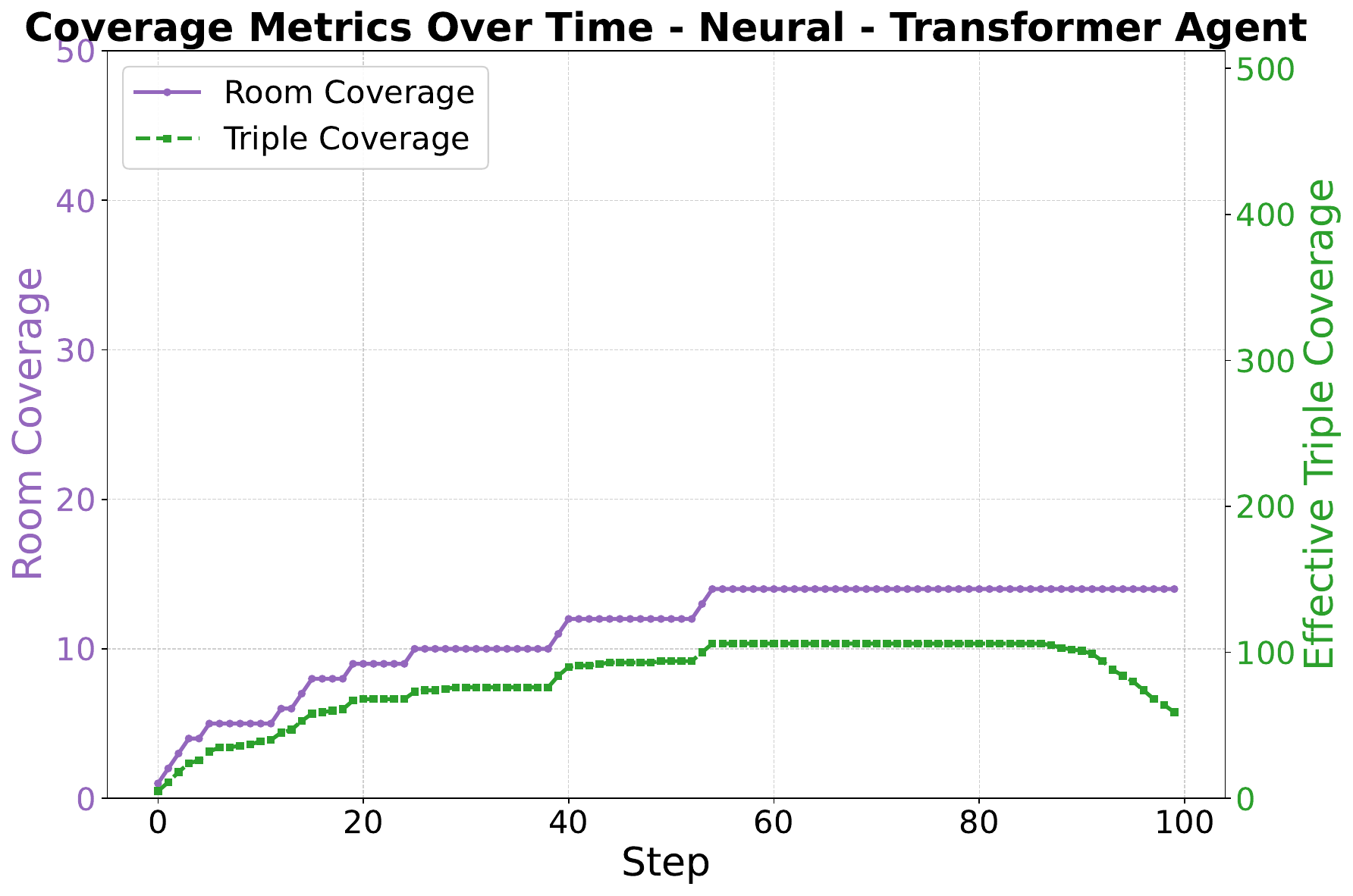}
        \textbf{(b) Neural -- Transformer}
    \end{minipage}

    \caption{Coverage metrics for the four agents, at the long-term memory capacity of 512. Neural agents halt exploration early, while symbolic agents visit all $49$ rooms and accumulate nearly all unique triples. Raw values can be found in the anonymous repository.}
    \label{fig:coverage}
\end{figure}

Figure~\ref{fig:coverage} shows, for the long-term memory capacity of $512$, the coverage metrics of the four agents: RDF, RDF-star, LSTM, and Transformer. Each subplot includes the cumulative number of unique rooms visited and the number of unique $(s,p,o)$ triples stored in memory at each timestep. Raw values are provided in the anonymous repository.

\paragraph{Symbolic agents.}
Both symbolic agents achieve full coverage of all $49$ rooms in the environment. Moreover, the RDF-star agent reaches full room coverage slightly earlier (timestep~70) than the RDF agent (timestep~74). The temporal qualifiers in RDF-star allow it to track time-varying wall configurations more accurately, yielding a more up-to-date inferred map of the environment and enabling more efficient exploration. Triple-coverage results follow the same pattern: symbolic agents accumulate nearly complete internal maps because they continue to explore until all reachable structure has been observed.

\paragraph{Neural agents.}
In contrast, the LSTM and Transformer agents stop increasing room coverage long before the end of the episode (around step~50 for the Transformer and step~30 for the LSTM). A plausible explanation is that neural policies must couple exploration and answering within a single high-cardinality action space. Once training converges to a locally rewarding but suboptimal behavior—such as repeatedly answering queries with moderate reward or oscillating among familiar states—the agents no longer discover new rooms. Because the reward signal is sparse and not explicitly tied to exploration, the policies collapse into low-entropy, non-exploratory behaviors. These plateauing exploration patterns also explain the limited triple coverage: neural agents simply encounter fewer states from which to build internal structure.

\subsection{Memory-State Evolution (RDF-Star)}

\begin{figure}[H]
    \centering
    \includegraphics[width=0.4\linewidth]{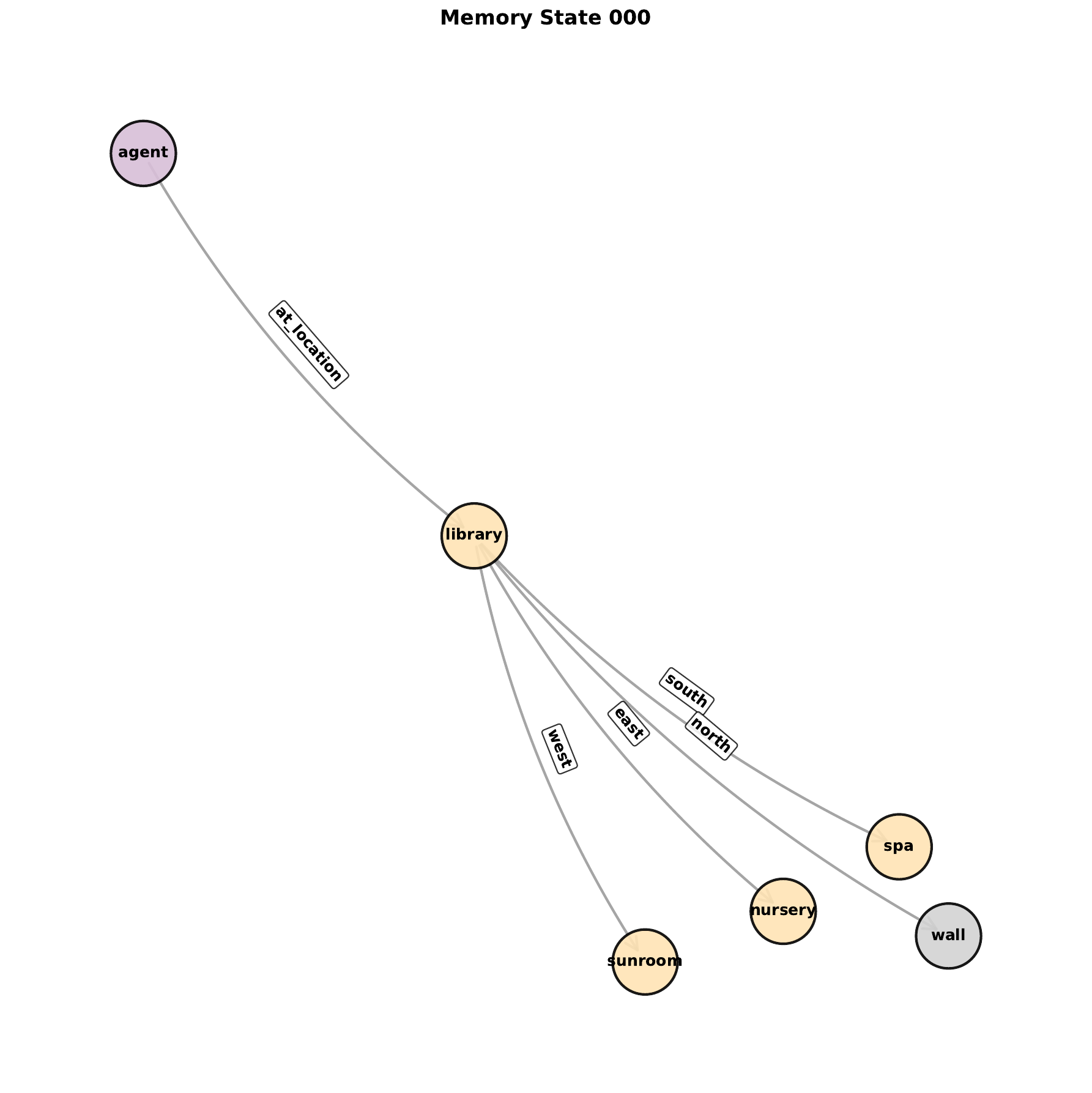}
    \hfill
    \includegraphics[width=0.5\linewidth]{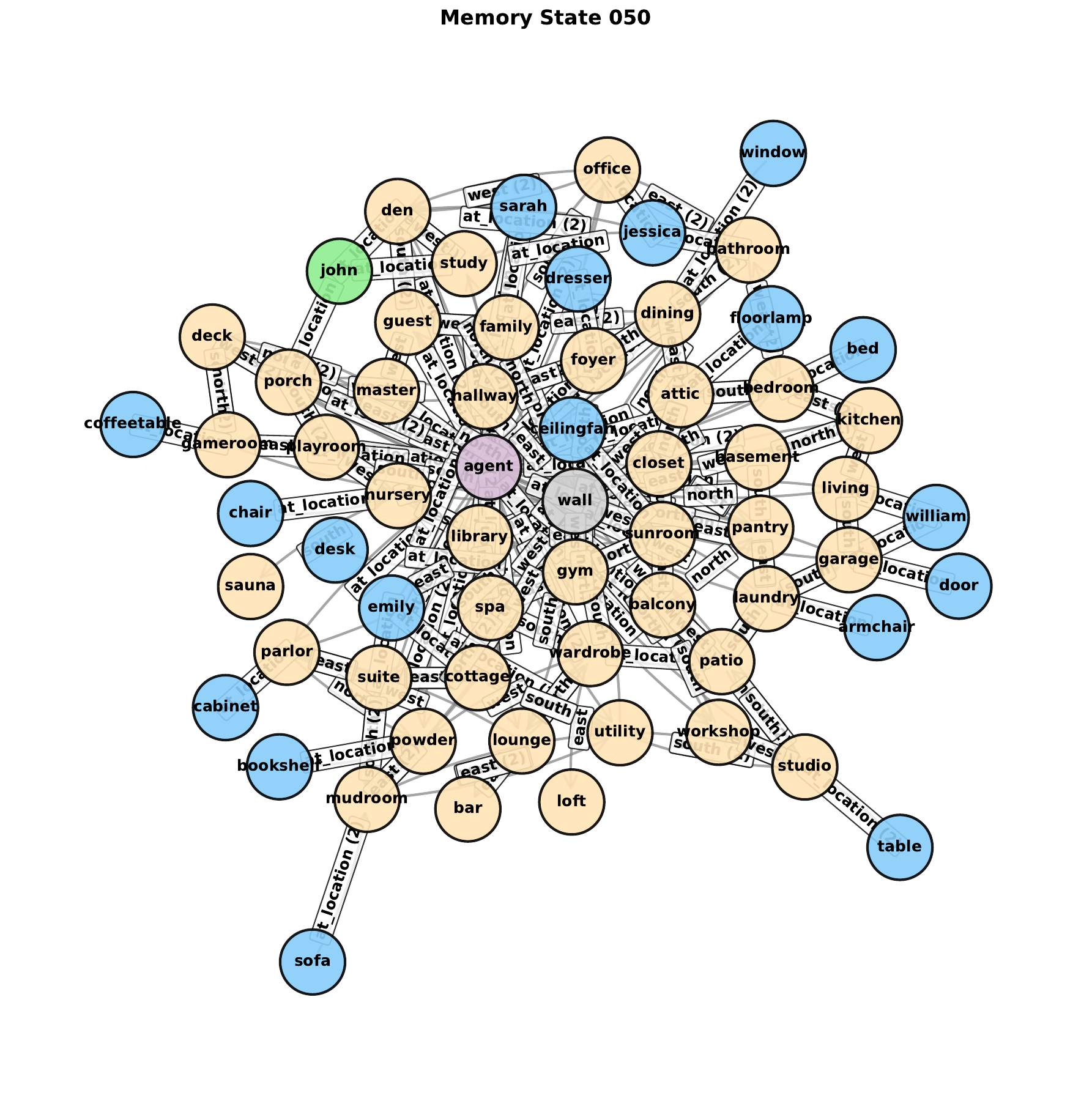}

    \vspace{0.6em}

    \includegraphics[width=0.7\linewidth]{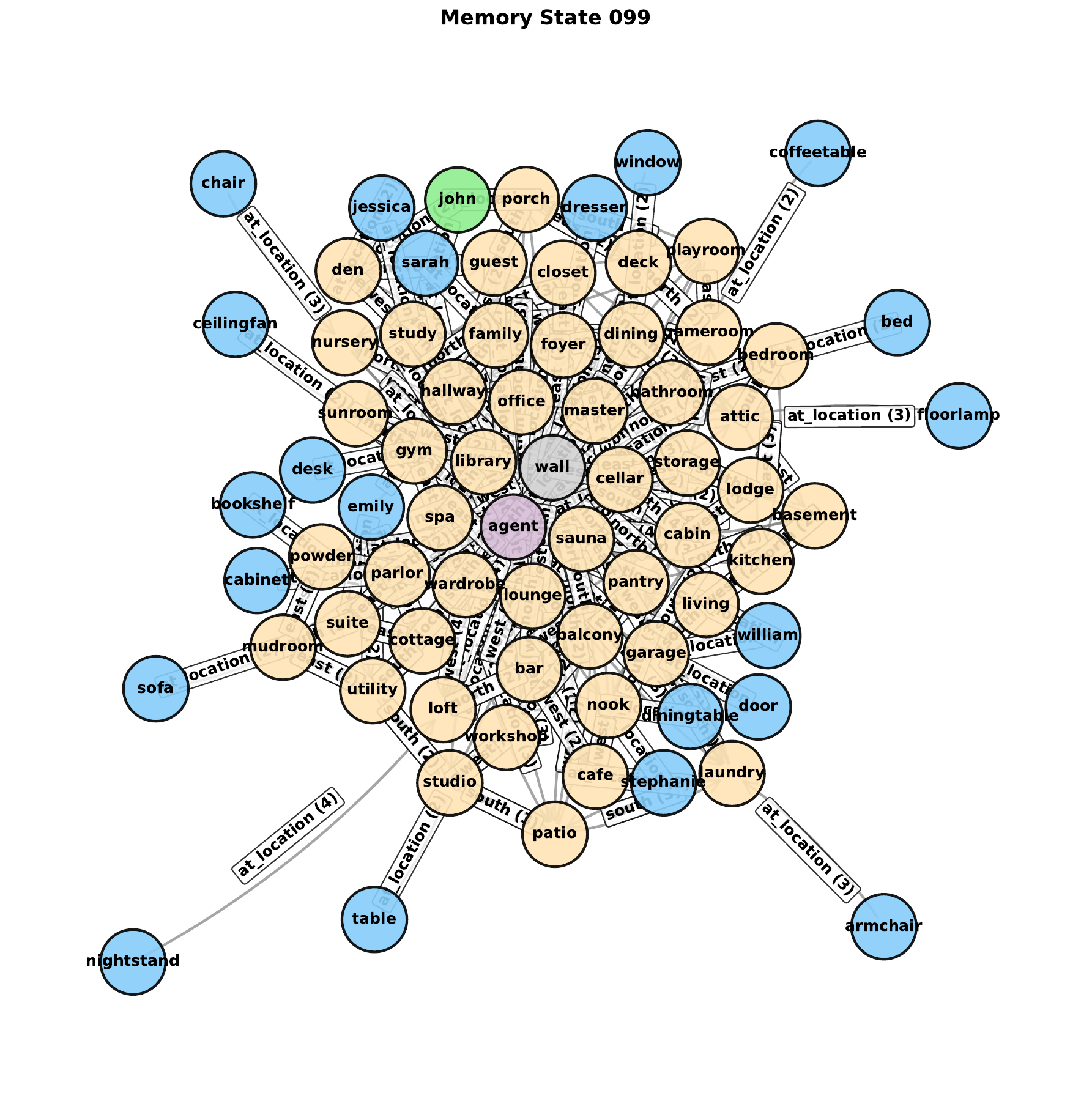}
    \caption{Memory state of the RDF-star agent at $t=0$, $t=50$, and $t=99$ (capacity $512$).  Counts in parentheses (e.g., ``\texttt{at\_location (3)}'') indicate how many RDF-star memories share the same main triple but differ in their temporal qualifiers.}
    \label{fig:memory_states}
\end{figure}

To illustrate the expressive advantage of temporal symbolic memory, Figure~\ref{fig:memory_states} shows the RDF-star agent’s long-term memory at timesteps $t=0$, $t=50$, and $t=99$, using capacity $512$. Each edge label includes the number of associated memories when more than one quadruple exists. Node colors correspond to semantic categories (rooms, static objects, moving objects, walls, and agent).  

The graphs show the internal map becoming increasingly complete over time. Static objects (blue) appear prominently because the agent’s deterministic BFS traversal guarantees that every room is visited at least once, and static objects never move; each one is therefore observed reliably when its room is first encountered. In contrast, moving objects (green) may leave a room before the agent arrives, their trajectories may intersect the traversal path only rarely, and their past locations may later be evicted under capacity pressure. As a result, moving objects are represented by far fewer stored triples—sometimes only a single observation—despite being present throughout the hidden state.

Full memory evolution images for all agents (from $t=0$ to $t=99$) are available in the anonymous repository.

\section{Related Work}
\label{sec:related_work}

\paragraph{Temporal and hyper-relational knowledge graphs.}
The Semantic Web stack provides a mature basis for representing structured data as
RDF graphs and querying them with SPARQL~\cite{Seaborne:13:SQL}.
RDF-star and SPARQL-star extend this model with \emph{statements about statements},
allowing triples such as $\ll s\ p\ o\gg$ to be annotated with additional facts and
queried directly~\cite{hartig2021foundationsalternativeapproachreification}.
Temporal and hyper-relational knowledge graphs build on these ideas by attaching
time and other qualifiers to edges and have been extensively studied for tasks such
as link prediction and event forecasting
(e.g.,~\cite{DBLP:journals/corr/abs-1809-03202}).
In contrast, we use a temporal KG not primarily as a prediction target but as the
\emph{internal state} of an agent: every observation becomes a reified RDF-star
statement with temporal qualifiers, and downstream decisions (answering and exploration)
are implemented as queries and updates over this KG.

\paragraph{Memory under partial observability.}
In partially observable environments, agents must aggregate information over time.
A large body of work uses neural memories to maintain an internal state proxy, including
recurrent architectures, memory-augmented networks such as Neural Turing Machines
and Differentiable Neural Computers~\cite{graves2014neural,graves2016hybrid},
and long-context models with compression or retrieval
(e.g., MERLIN and related world-model agents~\cite{Wayne2018MERLIN,DBLP:journals/corr/abs-1803-10122}).
These approaches typically store history in dense vectors or key–value tensors,
which makes it difficult to inspect what exactly is remembered at a given time.
Our work instead treats the agent’s long-term memory as an explicit temporal
knowledge graph in RDF-star: each stored fact is a verifiable triple with qualifiers,
and the effect of capacity limits or update rules on question answering can be
inspected directly at the graph level. Alongside this, we include simple neural
baselines that operate on observation histories without explicit KG structure,
highlighting the behavioural differences between latent and symbolic memories~\cite{kimura2021neurosymbolicreinforcementlearningfirstorder}.

\paragraph{Graph-structured environments and KG-based agents.}
Several interactive environments expose graph-structured or relational state~\cite{chaplot2020neuraltopologicalslamvisual}, and neural architectures explicitly designed for graph-structured reasoning~\cite{yun2020graphtransformernetworks}.
TextWorld and Jericho provide text-based games where an evolving symbolic world
model can be extracted and used for control, leading to agents that query or learn
over dynamic knowledge graphs~\cite{Cote2018TextWorld,Hausknecht2020Jericho,Ammanabrolu2020Graph,oh2017valuepredictionnetwork}. Agents with multiple KG-based memory stores have also been explored~\cite{10.1609/aaai.v37i1.25075}. Other systems use KGs as world models for planning, question answering, or recommendation~\cite{Hogan_2021,ehrlinger2016towards,bordes2015largescalesimplequestionanswering,yih-etal-2015-semantic}. Compared to these settings, our focus is deliberately narrow: we design the Room Environment v3 where (i) the hidden state and observations are \emph{already} KGs, (ii) the question format is fixed, and (iii) the environment parameters (number of rooms, walls, objects, episode length) are easily adjusted. This allows us to isolate the role of temporal KG memory and to compare symbolic and neural baselines under controlled conditions rather than to optimize a full
task pipeline.

\paragraph{Semantic Web, agents, and explainability.}
Within the Semantic Web community, KGs are increasingly used to support
decision-making and explainable reasoning in multi-step processes, e.g., planning, data analytics, and interactive decision support.
Traceability and provenance are central: KG representations allow one to link
conclusions back to the underlying facts~\cite{10.1145/775152.775249}. Our design follows this line by making the agent’s long-term memory itself a temporal KG; question answering is implemented as SPARQL-star-style retrieval over that memory, and memory contents can be
visualized as graphs over time. This provides a compact testbed for studying how
temporal qualifiers (e.g., recency and recall counts) interact with partial
observability and capacity limits, and how symbolic KG memories compare to simple neural baselines in terms of generalization from training to held-out question orders.

\section{Conclusion}

We presented the Room Environment v3 in which both the hidden state and the agent’s memory are represented as knowledge graphs, enabling a controlled study of temporal memory under partial observability. Using a lightweight RDF-star-based memory with temporal qualifiers, symbolic agents achieved full environment coverage and substantially higher QA accuracy than neural sequence baselines under the same environment and query conditions. The results highlight the effectiveness and interpretability of temporal knowledge graphs as agent memory, and provide a compact testbed for future work on semantic representations, temporal update rules, and neuro-symbolic memory models.

\bibliographystyle{plainnat}
\bibliography{references}

\end{document}